\documentclass[fleqn]{SCYE}
\usepackage{CJKutf8}
\usepackage[numbers]{natbib}
\usepackage{setspace}
\usepackage{multirow}

\def\newcite#1{\citeauthor{#1} (\citeyear{#1}) \cite{#1}}
\def\workcite#1{(\citeauthor{#1}, \citeyear{#1}) \cite{#1}}
\def\sworkcite#1{\citeauthor{#1} (\citeyear{#1}) \cite{#1}}

\begin{document}
\begin{CJK}{UTF8}{gbsn}


\title{A Survey of Syntactic-Semantic Parsing Based on Constituent and Dependency Structures}
{A Survey of Syntactic-Semantic Parsing }

\author{Meishan Zhang}{mason.zms@gmail.com}%



\address{School of New Media and Communication, Tianjin University, China}




\maketitle



\begin{multicols}{2}

\section*{Abstract}
Syntactic and semantic parsing has been investigated for decades,
which is one primary topic in the natural language processing community.
This article aims for a brief survey on this topic.
The parsing community includes many tasks,
which are difficult to be covered fully.
Here we focus on two of the most popular formalizations of parsing:
constituent parsing and dependency parsing.
Constituent parsing is majorly targeted to syntactic analysis,
and dependency parsing can handle both syntactic and semantic analysis.
This article briefly reviews the representative models of constituent parsing and dependency parsing,
and also dependency graph parsing with rich semantics.
Besides, we also review the closely-related topics such as cross-domain, cross-lingual and joint parsing models,
parser application as well as corpus development of parsing in the article.

\section{Introduction}\label{section1}
Sentence-level syntactic and semantic parsing is one major topic in the natural language processing (NLP) community,
which aims to uncover the internal structural relations in sentences \cite{manning1999foundations,kubler2009dependency,zcq2013,jurafsky2019speech}.
From the view of linguistics,
the goal of parsing is to disclose how words are combined to form sentences and the rules that govern the formation of sentences.
On the other hand, from the view of NLP applications,
parsing can be beneficial for a number of tasks,
such as machine translation, question answering, information extraction, sentiment analysis and generation \cite{yamada-knight-2001-syntax,chan-roth-2011-exploiting,zou2015sentiment},
and the performance of parsing matters greatly.

Parsing has been extensively studied for decades.
The goal of syntactic parsing is to derive the syntax information in sentences,
such as the subjects, objects, modifiers and topics.
There have been a number of achievements for the task,
and large-scale corpora for a range of languages have been already available.
Compared with syntactic parsing,
semantic parsing is much more difficult due to the complex structure of various semantics such as predicate-argument,
and it is also a long-range goal of NLP.
With the recent advance in data-driven machine learning models,
semantic parsing has received increasing interests, especially under the neural setting.
Several datasets based on certain formalizations have been developed to facilitate research.

Parsing often relies on specific grammars,
which are used to refine the output structures of syntax and semantics.
There are many sophisticated grammars for accurately expressing the syntactic and semantic information at the sentence-level.
In this paper, we focus on two popular grammars which are concerned mostly.
Context-free grammar (CFG), well known as \textbf{constituent parsing} (or phrase-structure parsing) \cite{jurafsky2019speech}
(thus, also as constituent grammar or phrase-structure grammar),  adopts hierarchal phrase-structural trees to organize sentence-level syntactic information,
which has been researched intensively since very early.
Dependency grammar is another widely-adopted grammar for syntactic and semantic parsing,
where words are directly connected by dependency links, with labels indicating their syntactic or semantic relations \cite{kubler2009dependency}.
Because of the conciseness and easy annotation of dependency structures,
\textbf{dependency parsing} has received more attention than constituent parsing.

Besides, there are many other great grammars.
The representative topics include combinatory categorial grammar (CCG),
head-driven phrase structure grammar (HPSG),
lexical functional grammar (LFG),
abstract meaning representation (AMR),
minimal recursion semantics (MRS),
universal conceptual cognitive annotation (UCCA)
and also several logic-targeted formalizations.
All these categories have been researched for a long time
and in particular several of which are now quickly developed because of the powerfulness of neural networks
as well as pretrained contextualized word representations.
However, this article leaves these studies for future more comprehensive surveys.

\setlength{\tabcolsep}{4.0pt}
\begin{table}[H]
\begin{threeparttable}
\footnotesize
\caption{A comparison of representative constituent parsing models, where phrase-level F1 scores are reported, PTB and CTB are two benchmark datasets for the English and Chinese parsing, respectively.  } \label{table:const:performance}
\begin{tabular}{l|c|cc}
\hline
Model &  Main Features &  PTB &   CTB \\ \hline \hline
\multicolumn{4}{c}{ \texttt{Chart-based, Statistical Models}}  \\ \hline
\sworkcite{collins-1997-three} &  head-lexicalization & 88.2  & N/A \\
\sworkcite{Charniak2000}  &  max-entropy  & 89.5  & 80.8 \\
\sworkcite{mcclosky-etal-2006-effective} &  self-training  & \bf 92.3  & N/A \\
\sworkcite{petrov-klein-2007-improved} &   PCFG  & 90.1 &  \bf 83.3 \\
\sworkcite{hall-etal-2014-less} &  CRF & 89.9  & N/A \\
\hline
\hline
\multicolumn{4}{c}{ \texttt{Transition-based, Statistical Models}}  \\ \hline
\sworkcite{sagae-lavie-2005-classifier}  &  greedy & 86.0  & N/A \\
\sworkcite{zhu-etal-2013-fast}   &  global learning, beam & \bf 91.3  & \bf 85.6 \\ \hline
\hline
\multicolumn{4}{c}{ \texttt{Chart-based, Neural Models}}  \\ \hline
\sworkcite{socher-etal-2013-parsing} & recursive NN & 90.4  & N/A \\
\sworkcite{durrett-klein-2015-neural} & CNN & 91.1  & N/A \\
\sworkcite{stern-etal-2017-minimal} &  LSTM, span  & 91.8 &  N/A \\
\sworkcite{kitaev-klein-2018-constituency} (a)  & self-attentive & 93.5  & N/A \\
\sworkcite{kitaev-klein-2018-constituency} (b)  & +ELMo & \bf 95.1  & N/A \\ \hline
\hline
\multicolumn{4}{c}{ \texttt{Transition-based, Neural Models}}  \\ \hline
\sworkcite{wang-etal-2015-feature}  &  neural+discrete & 90.7  & \bf 86.6 \\
\sworkcite{watanabe-sumita-2015-transition}  &  global learning, beam & 90.7  & N/A \\
\sworkcite{dyer-etal-2016-recurrent} & language modelling &  92.4  & 82.7 \\
\sworkcite{cross-huang-2016-span}  &   dynamic oracle & 91.3  & N/A \\
\sworkcite{liu-zhang-2017-order} &   in-order & 91.8  & 86.1 \\
\sworkcite{fried-klein-2018-policy}  &   policy gradient &  92.6  & 86.0 \\
\sworkcite{kitaev2019tetra}  &   policy gradient & \bf 95.4  & 86.0 \\
\hline
\hline
\multicolumn{4}{c}{ \texttt{Other Methods (report neural models only) }}  \\ \hline
\sworkcite{shen-etal-2018-straight}  & distance to tree & 91.8  & 86.5 \\
\sworkcite{teng-zhang-2018-two}  & local classification & 92.7  & 87.3 \\
\sworkcite{vilares-etal-2019-better}  & sequence labeling & 91.1  & 85.6 \\
\sworkcite{zhou-zhao-2019-head} & HPSG grammar & \bf 96.3  & \bf 92.2 \\
\sworkcite{mrini2019rethinking} & HPSG, improved attention  & \bf 96.3  & \bf N/A \\
\hline
\end{tabular}
\end{threeparttable}
\end{table}

Here we make a brief survey for syntactic and semantic parsing based on \textbf{constituent grammar} and \textbf{bi-lexicalized dependency grammar}.
In Section 2 and 3 we review the studies of constituent parsing and dependency parsing, respectively,
where the dependency parsing is based tree structure and specifically targeted to syntax.
We further investigate semantic-oriented dependency graph parsing in Section 4.
Section 5 and 6 review cross-domain and cross-lingual parsing, which is one hot direction.
Section 7 reviews the joint models which are targeted to parsing as the final goal,
while Section 8 reviews the parser application strategies, where parsers are evaluated on downstream applications.
Section 9 introduces the related treebank work, which serves the major training corpus for various parsers as well as for parser model evaluations.
Finally, in Section 10, the conclusion and future work are summarized.

\section{Constituent Parsing}
\begin{figure}[H]
\includegraphics[scale=0.6]{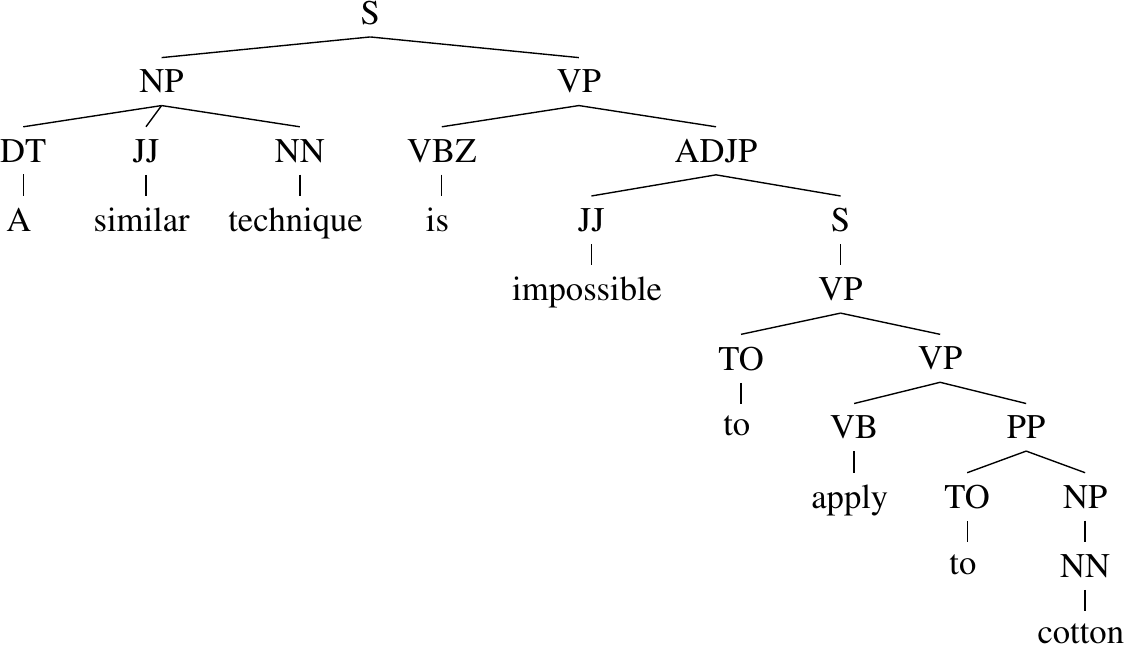}
\caption{An example of constituent tree.}
\label{fig-constituent}
\end{figure}

Constituent parsing is one fundamental task for syntax parsing,
which has received great interest for decades \cite{manning1999foundations,zcq2013,jurafsky2019speech}.
Figure \ref{fig-constituent} shows an example constituent tree,
where nodes in the constituent tree are constituent spans, also known as phrases.
The goal of constituent parsing is to uncover these phrases as well as their relations.
The standard evaluation method of constituent parsers is based on recognition of the phrases,
where precision, recall and the F1-measure scores are adopted as the major metrics.

The mainstream approaches of constituent parsing include the chart-based and the transition-based models.
Current neural models have achieved state-of-the-art performances under both two kinds of methods.
In fact, neural constituent parsing starts very early before the prosperity of deep learning \cite{henderson-2004-discriminative} .
In this section, first we introduce the chart-based and transition-based constituent models,
and then show several other models out of the two categories.
Here before the detailed introduction,
we show an overall picture of the performances of various representative constituent parsers in Table \ref{table:const:performance},
where ensemble models are excluded for fair comparisons.

\subsection{Chart-Based Parsing}

\subsubsection{Statistical Models}
Early successful constituent parsing models exploit the productive CFG rules to guide the generation of constituent trees.
The chart parsing algorithms are exploited universally for decoding,
and most of the effort is focused on the refinement of CFG rules, which serve as the major sources of parameter estimation.
\newcite{collins-1997-three} and \newcite{Charniak2000} extend probabilistic context-free grammar (PCFG) with head lexicalization,
associating PCFG rules with head words, which can effectively boost the PCFG parsing performance.
Unlexicalized models have also received great attention,
by using fine-grained structural annotation \cite{klein-manning-2003-accurate} or automatic latent variables
\cite{petrov-klein-2007-improved} to enrich PCFG rules,
leading to comparable or even better performance than lexicalized models.

The above models suffer the difficulty of integrating non-local features
since future decisions are invisible during decoding which is critical for global inference.
Condition random field (CRF) is one way for global modeling.
\newcite{hall-etal-2014-less} propose a strong constituent parsing model by adapting the standard n-gram CRF models for CFG,
and meanwhile presenting rich sophisticated features.
The dependencies among adjacent CFG rules can be modeled,
which are used for global inference.

\subsubsection{Neural Models}
\newcite{socher2010learning} is the first work to define scores over phrases by recursive neural networks.
The CFG-based constituent trees can be naturally modeled in this way.
Neural CRF parsing is accordingly proposed by \newcite{durrett-klein-2015-neural},
which can be regarded as a neural enhancing of \newcite{hall-etal-2014-less}.
The work simply uses feed-forward neural networks to encode atomic features instead of human composition.
Notice that it is different from \newcite{socher2010learning} as no recursive composition is used here.

\newcite{stern-etal-2017-minimal} propose state-of-the-art chart-based neural models.
On the one hand, they use deep bidirectional long-short term memory (LSTM) neural networks to enhance sentence representations,
designing sophisticated strategies for span representation.
On the other hand, they also adopt top-down incremental parsing for decoding,
which dilutes the differences between chart-based and transition-based approaches.
Their results are very strong on par with the state-of-the-art transition-based methods at the same time.
The work is further followed by \newcite{gaddy-etal-2018-whats} with extensive analysis and \newcite{kitaev-klein-2018-constituency} with a self-attentive encoder.
In particular, \newcite{kitaev-klein-2018-constituency} exploit contextualized word representation including ELMo \cite{peters-etal-2018-deep}
and BERT \cite{devlin-etal-2019-bert},
leading to almost the best parsing performance in the literature.

\subsection{Transition-Based Parsing}

\subsubsection{Statistical Models}
The transition-based models demonstrate highly promising for constituent parsing \cite{ratnaparkhi-1997-linear,sagae-lavie-2005-classifier}.
The key idea is to define a transition system with transition states and actions,
where states denote partial parsing outputs,
and actions specify next-step state-transition operations.
Transition actions indicate the incremental tree construction process.
For constituent parsing, typical actions include the \emph{shift} to building terminal nodes, the \emph{unary} to building unary nodes,
and the \emph{binary} to building binary nodes.
The details can be referred to as \newcite{sagae-lavie-2005-classifier}.
The model is also commonly referred to as the shift-reduce model, where \emph{unary} and \emph{binary} are actions of reduction.
By converting constituent parsing into predicting a sequence of transition actions,
discriminant classifiers such as max-entropy and support vector machine (SVM) can be applied for the prediction,
with rich manually-crafted features.

The initial shift-reduce model classifies the sequence of actions for a single constituent tree separately,
greedily searching for the best output constituent tree,
which may suffer the error propagation problem since the early step errors can affect later predictions.
To this end, globally modeling with beam search is proposed to alleviate the problem,
which decodes the total sequence of actions for a full constituent tree as a whole \cite{zhang-clark:2009:IWPT09,zhu-etal-2013-fast}.
The discriminative perceptron-style online learning greatly promotes this line of work \cite{collins:2002:EMNLP02},
which enables legal parameter optimizations towards inexact search.
For feature engineering, the contextual lexicalized words, POS tags, distances and their compositions are all extensively investigated,
which can be referred to \cite{zhu-etal-2013-fast} for details.

\subsubsection{Neural Models}

\newcite{watanabe-sumita-2015-transition} and \newcite{wang-etal-2015-feature} could be the direct extensions of \newcite{zhu-etal-2013-fast} by using neural networks.
The composition of atomic features is achieved by feed-forward neural networks.
\newcite{cross-huang-2016-incremental} find that the greedy style decoding can also achieve highly competitive performance when a deep LSTM encoder is exploited.
Then, several studies suggest dynamic oracles to optimize greedy constituent parsers
\cite{cross-huang-2016-span,coavoux-crabbe-2016-neural}.
The main idea is to let models make optimum decisions when facing erroneous transition states \cite{goldberg-nivre-2012-dynamic}.
A proportion of training instances with erroneous transition states and their oracle actions are added into the original training corpus.

There have been several studies exploiting different transition strategies.
\newcite{dyer-etal-2016-recurrent} suggest the recurrent neural work grammar (RNNG),
which is a top-down transition-based system.
\newcite{liu-zhang-2017-order} design an in-order transition system to make a compromise between top-down and bottom-up transitions.
\newcite{coavoux-etal-2019-unlexicalized} present a novel system with an additional GAP action for discontinuous constituency parsing,
and they also find that unlexicalized models give better performance.
\newcite{fernandez2019faster} optimize the transition actions to facilitate the construction of non-binarized constituent nodes,
avoiding the preprocessing of binarization for constituent trees.
\newcite{kitaev2019tetra} suggest the tetra-tagging system, which combines sequence tagging and transition action classification.
The system achieves state-of-the-art performance on the benchmark PTB dataset with BERT representations.

\subsection{Other Frameworks}

Neural networks such as deep LSTM and multi-head self-attention
are capable of encoding global features implicitly into their final representations,
which weakens the role of decoding as a source of feature induction.
Based on the observation,
several studies attempt to use simple frameworks,
aiming for a wide community for parsing.

One representative attempt is to exploit neural sequence-to-sequence models for structural constituent parsing \cite{vinyals2015grammar,choe-charniak-2016-parsing}.
The key idea is to first linearize a phrase-structural constituent tree into a sequence of symbols by certain traversing strategies,
and then directly feed the pair of input words and output symbols into a standard sequence-to-sequence model.
These models require large-scale corpora for training,
which could be obtained by auto-parsed high-confidence constituent trees from other state-of-the-art parsers.

Neural sequence labeling models have also been investigated for constituent parsing \cite{gomez-rodriguez-vilares-2018-constituent}.
\newcite{gomez-rodriguez-vilares-2018-constituent} propose the first work of this line,
which exploits the lowest common ancestor between adjacent words as clues to encode the word roles.
\newcite{vilares2020parsing} extend the work by language modeling and enhance parsing with pretraining.
Further, more direct schemes have been proposed with local modeling for constituent parsing.
\newcite{shen-etal-2018-straight} directly predict the distance of constituent phrases
and then decode greedily in a top-down way for a full constituent tree.
Similarly, \newcite{teng-zhang-2018-two} propose two models based on local span prediction,
achieving highly competitive performance on par with transition-based models.
Recently, \newcite{zhou-zhao-2019-head} present to exploit the HPSG-based grammar for constituent parsing,
and further power the model with XLNet word representations \cite{yang2019xlnet},
achieving the top performances for both CTB and PTB datasets.
\newcite{mrini2019rethinking} revise the multi-head self-attention mechanism in \newcite{zhou-zhao-2019-head},
leading to a similar performance with a smaller number of layers.

\setlength{\tabcolsep}{1.5pt}
\begin{table}[H]
\begin{threeparttable}
\footnotesize
\caption{A comparison of representative dependency parsing models, where UAS are reported, PTB and CTB5.1 (CTB in the Table for short) are two benchmark datasets for the English and Chinese parsing, respectively.  } \label{table:dep:performance}
\begin{tabular}{l|c|cc}
\hline
Model &  Main Features &  PTB &   CTB \\ \hline \hline
\multicolumn{4}{c}{ \texttt{Graph-based, Statistical Models}}  \\ \hline
\sworkcite{mcdonald-etal-2005-online} &  1-order & 90.9  & 83.0 \\
\sworkcite{McDonald2006}  &   2-order & 91.5  & 85.2 \\
\sworkcite{koo-etal-2008-simple} &   word clusters  & 93.2 &  N/A \\
\sworkcite{chen-etal-2009-improving} &  auto subtrees & 93.2 &  86.7 \\
\sworkcite{bohnet-2010-top} & feature hashing & 92.9 & N/A  \\
\sworkcite{koo-collins-2010-efficient}  &   3-order & 93.0  & 86.0 \\
\sworkcite{ma-zhao-2012-fourth}  &  4-order & \bf 93.4  & \bf 87.4 \\
\hline
\hline
\multicolumn{4}{c}{ \texttt{Transition-based, Statistical Models}}  \\ \hline
\sworkcite{nivre-cl08} (a) &  arc-standard & 89.7  & 82.7 \\
\sworkcite{nivre-cl08} (b)  &  arc-eager & 89.9  & 80.3 \\
\sworkcite{zhang-clark-2008-tale}  &   global learning, beam & 91.4  & 84.3 \\
\sworkcite{zhang-nivre-2011-transition}  &   rich non-local features & \bf 92.9  & \bf 86.0 \\
\sworkcite{goldberg-nivre-2012-dynamic}  & dynamic oracle & 91.0  & 84.7 \\ \hline
\hline
\multicolumn{4}{c}{ \texttt{Graph-based, Neural Models}}  \\ \hline
\sworkcite{pei-etal-2015-effective} & feed-forward & 93.3  & N/A \\
\sworkcite{zhang-etal-2016-probabilistic} & CNN & 93.4  & 87.7 \\
\sworkcite{wang-chang-2016-graph} &  2-layer LSTM  & 94.1 &  87.6 \\
\sworkcite{kiperwasser-goldberg-2016-simple}  & 2-layer LSTM & 93.1  & 86.6 \\
\sworkcite{dozat2016deep}  & 3-layer LSTM, biaffine  & 95.7  & 88.9 \\
\sworkcite{li2019self} (a) & self-attentive & 95.9  & 92.2 \\
\sworkcite{li2019self} (b)  &  +ELMO & 96.6  & 90.3 \\
\sworkcite{li2019self} (c)  &  +BERT & \bf 96.7  & \bf 92.2 \\
\sworkcite{ji-etal-2019-graph}  &  GNN & 96.0  & N/A \\ \hline
\hline
\multicolumn{4}{c}{ \texttt{Transition-based, Neural Models}}  \\ \hline
\sworkcite{chen-manning-2014-fast}  &  feed-forward & 91.8  & 83.9 \\
\sworkcite{dyer-etal-2015-transition}  &  stack-LSTM & 93.1  & 87.2 \\
\sworkcite{zhou-etal-2015-neural}  &   global learning, beam & 93.3  & N/A \\
\sworkcite{andor-etal-2016-globally}  &   global learning, beam & 94.6  & N/A \\
\sworkcite{kiperwasser-goldberg-2016-simple} & 2-layer LSTM &  93.9  &  87.6 \\
\sworkcite{ballesteros-etal-2017-greedy}  & char, stack-LSTM & 93.6  & 87.6 \\
\sworkcite{ma-etal-2018-stack} &  3-layer LSTM  & \bf 95.9  & \bf 90.6   \\  \hline
\hline
\multicolumn{4}{c}{ \texttt{Other Methods (report neural models only) }}  \\ \hline
\sworkcite{kiperwasser-goldberg-2016-easy}  & easy-first & 93.0  & 87.1 \\
\sworkcite{li-etal-2018-seq2seq}  & sequence-to-sequence & 92.1  & 86.2 \\
\sworkcite{strzyz-etal-2019-viable}  & sequence labeling & 93.7  & N/A \\
\sworkcite{zhou-zhao-2019-head} & HPSG grammar & 97.2  & \bf 91.2 \\
\sworkcite{mrini2019rethinking} & HPSG, improved attention  & \bf 97.3  & \bf N/A \\
\hline
\end{tabular}
\end{threeparttable}
\end{table}

\subsection{Semi-Supervised Models}
The semi-supervised architecture aims to enhance a supervised model by statistical information extracted from raw text.
\newcite{mcclosky-etal-2006-effective} present the first work which achieves improved performance for constituent parsing by self-training,
and \newcite{mcclosky-etal-2008-self} study self-training empirically to show the conditions of usefulness.
\newcite{candito2009improving} exploit unsupervised word clusters learned from raw text to enhance constituent parsing.
While recent studies shift to the neural network setting, the borderline between semi-supervised and supervised is becoming vague,
as pretraining from raw text is one critical for the successfulness of neural models.

\subsection{Model Ensemble}
The model ensemble is one effective way to boost the performance of constituent parsing.
Initial work focuses on the output reranking \cite{collins-koo-2005-discriminative,huang-2008-forest}.
We can take either the k-best outputs of a constituent parser or one-best outputs from heterogeneous parsers as the inputs,
and then build a new constituent tree by using a feature-rich reranking model.
Benefiting from sophisticated manually-crafted non-local features,
the framework can improve the parser performance significantly.
However, related studies under the neural setting have received much less concern,
which can be potentially due to that the majority of state-of-the-art neural models exploit the same sentence encoders,
indicating that features are resemble in different kinds of models, and meanwhile homogeneous ensemble (e.g., different random seeds)
by simply voting can achieve unsurpassable performances.

\begin{figure}[H]
\includegraphics[scale=0.75]{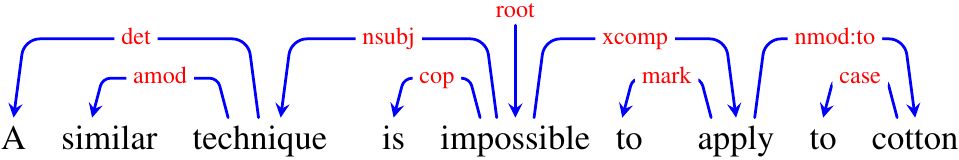}
\caption{An example of dependency tree.}
\label{fig-dependency}
\end{figure}

\section{Dependency Parsing}
Dependency parsing is developed for syntax and semantic analysis by using bilexicalized dependency grammar,
where all syntactic and semantic phenomena are represented by bilexicalized dependencies \cite{kubler2009dependency}.
Figure \ref{fig-dependency} shows an example tree of dependency parsing.
For the evaluation of various dependency parsers, dependency accuracy is used as the major metric,
in terms of the unlabeled attachment score (UAS) and the labeled attachment score (LAS).
In the early stage, dependency parsing is constrained to trees, projective or non-projective \cite{hajic-etal-2009-conll1}.
Recently, several studies have devoted to dependency parsing over graphs \cite{oepen2015semeval}.
On the one hand, initial dependency trees are mostly syntactic oriented,
while recently there are growing interests focusing on semantic relations between words \cite{hajic-etal-2009-conll1,oepen2015semeval}.
This section mainly focuses on dependency tree parsing,
while dependency graph parsing will be discussed in the next section.

The majority of dependency parsing models can be divided into two types,
graph-based and transition-based \cite{nivre-mcdonald-2008-integrating},
both of which have been extensively investigated under the traditional statistical setting \cite{mcdonald-etal-2005-online,XavierCarreras2007,nivre-iwpt03,Yamada2003} and
the neural setting \cite{nivre-mcdonald-2008-integrating}.
There also exist other interesting approaches for dependency parsing outside the two categories \cite{li-etal-2018-seq2seq}.
Table \ref{table:dep:performance} shows an overall picture of the performances of several representative dependency parsers,
and all ensemble models are excluded in this table.
The graph-based and transition-based models are almost comparable (graph-based models are slightly higher) across both traditional statistical and neural settings,
and other types of models achieve good performance with the support of sophisticated neural networks.
Currently， neural models achieve state-of-the-art performances  for dependency parsing \cite{kulmizev-etal-2019-deep}.

\subsection{Graph-Based Parsing}

\subsubsection{Statistical Models}

Graph-based methods exploit the maximum spanning tree (MST) algorithm for decoding,
which decomposes a full dependency tree into small factors, such as dependency edges,
and scores the full tree by summing the scores of all the included factors.
The score of each factor can be calculated independently by the features extracted from it.
The models by using dependency edges as the basic scoring factor are referred to as first-order models,
where the order indicates the maximum number of edges in a factor.
\newcite{mcdonald-etal-2005-online} propose a feature-rich first-order MST parser based on discriminative max-margin training.

Later, higher-order MST parsers have been studied.
With larger factors, the parsing models can exploit more sophisticated features, and thus can potentially bring improved performance.
Second order MST parsing models have studied extensively \cite{mcdonald-06-phd-thesis,McDonald2006,XavierCarreras2007,Bohnet2010},
where the newly added features include the relations from parent-sibling and parent-child-grandchild factors.
Notice that higher-order MST decoding can have higher time complexity (i.e., from $O(n^3)$ to $O(n^4)$),
which may lead to intolerable parsing speed.
The problem could be largely alleviated by \newcite{Bohnet2010} with feature hashing.
\newcite{koo-collins-2010-efficient} propose an efficient third-order dependency parsing model,
which adds grand-sibling and tri-sibling features into the model.
\newcite{lei-etal-2014-low} exploit low-rank tensor to alleviate the burden of feature engineering.
Fourth-order dependency parsing has been investigated by \newcite{ma-zhao-2012-fourth}.
As a whole, second-order and third-order parsers could be good choices considering both performance and efficiency.

\subsubsection{Neural Models}
\newcite{pei-etal-2015-effective} present a graph-based neural model
by embedding all discrete atomic features in the traditional statistical MST models and
then composing these embeddings with a similar feed-forward network of \workcite{chen-manning-2014-fast}.
Convolution neural network is then applied for neural feature composition in \newcite{zhang-etal-2016-probabilistic}.
Following, deep bidirectional LSTMs are exploited to substitute the simple neural feed-forward network \cite{wang-chang-2016-graph,kiperwasser-goldberg-2016-simple}.
As sentence-level global information can be encoded through these neural structures,
the performance gap between first- and higher-order decoding is largely reduced.

\newcite{dozat2016deep} propose a deep biaffine parser which achieves the impressive performances,
boosting the UAS and LAS numbers into a new degree.
The parser exploits a three-layer bidirectional LSTM as the encoder,
and a biaffine operation as the decoder to score all possible dependency edges.
This work adopts several tricks to reach their final performance,
e.g., the node-level dropouts, and the same dropout mask at every recurrent timestep.
\newcite{li2019self} further enhance the biaffine parser with self-attentive encoder and
contextualized word representations such as ELMo and BERT \cite{peters-etal-2018-deep,devlin-etal-2019-bert}.
\newcite{ji-etal-2019-graph} exploit graph neural networks to better the input sentence encoder.

\subsection{Transition-Based Parsing}

Transition-based models have achieved great success on dependency parsing.
To some extent, the transition-based framework is then received great attention to other NLP tasks involving structural learning
because of the successfulness of dependency parsing.
For example, the transition-based constituent parsing is initially inspired by transition-based dependency parsing.
On the one hand, the transition-based models can obtain nearly equivalent performance compared with graph-based methods.
On the other hand, these models are highly efficient,  which can achieve linear time complexity.
Transition-based models convert dependency parsing into an incremental state-transition process,
where states denote partial outputs and they are advanced step by step by predefined transition actions.

\subsubsection{Statistical Models}

The initial work for feature-rich transition-based dependency parsing is suggested by \newcite{nivre-iwpt03} and \newcite{Yamada2003},
and then the framework is extensively investigated \cite{nivre-cl08,gomez-rodriguez-nivre-2013-divisible}.
There are two typical transition configurations, arc-standard and arc-eager, respectively,
which are comparable in parsing performances.
Typically, the transition actions include \emph{shift} operation (aiming for starting next word processing),
\emph{arc-left} (aiming for building a left directional dependency),
and \emph{arc-right} (aiming for right directional dependencies).
Besides, several researchers propose other transition configurations \cite{nivre-2009-non,sartorio-etal-2013-transition,gomez-rodriguez-nivre-2013-divisible,noji-miyao-2014-left},
which can handle various complex cases, such as non-projective dependencies.

Early transition-based methods usually exploit discriminative classifiers for action prediction when a certain transition state is given,
which processes the parsing in a local manner.
The scheme may suffer the error propagation problem, where early errors can influence future predictions.
To alleviate the problem, global learning with beam-search is one effective way.
\newcite{zhang-clark-2008-tale} firstly apply the strategy.
Rich global features that have been exploited in high-order graph-based dependency parsers can be also integrated
into the model \cite{zhang-nivre-2011-transition}.
The strategy can be also enhanced with dynamic programming further \cite{huang-sagae-2010-dynamic,kuhlmann-etal-2011-dynamic}.

Another alternative strategy is the dynamic oracle,
which is firstly proposed by \newcite{goldberg-nivre-2012-dynamic} for transition-based models by using arc-eager.
The method defines dynamic gold-standard oracle based on a sample of erroneous states,
and then add these instances to enhance model training.
Thus, we can minimize global performance losses when errors occur.
Although the strategy gives slightly worse performance than \newcite{zhang-nivre-2011-transition},
it enables dependency parsing in a greedy way, greatly increasing the parsing efficiency.
The strategy has been investigated by several studies with different configurations,
such as arc-standard and non-projective parsing \cite{goldberg-etal-2014-tabular,gomez-rodriguez-etal-2014-polynomial}.

\subsubsection{Neural Models}
\workcite{chen-manning-2014-fast} is one millstone work for neural dependency parsing,
which substitutes traditional manually-crafted discrete features with neural features.
The work uses simple feed-forward neural networks to compose the embeddings of all atomic features automatically,
and thus is free of feature engineering.
Finally, the proposed model obtained much better performance than the corresponding statistical baseline.
Pretrained word embeddings and the neural composition function are the keys to success.

There exist several directions to improve the performance of neural transition-based dependency parsing.
First, we can exploit better neural network structures.
Stack-LSTM is presented by \newcite{dyer-etal-2015-transition} and then followed by several studies \cite{ballesteros-etal-2015-improved,ballesteros-etal-2017-greedy,de-lhoneux-etal-2019-recursive},
which can represent transition states by utilizing partial structural information.
In parallel, deep bidirectional LSTM is also investigated \cite{kiperwasser-goldberg-2016-simple,ma-etal-2018-stack}.
\newcite{ma-etal-2018-stack} exploit a similar encoder as \newcite{dozat2016deep}, achieving slightly better performances than \workcite{dozat2016deep}.
In fact, with powerful neural encoders, especially pretrained contextualized word representations,
the performance difference between graph-based and transition-based is very marginal \cite{kulmizev-etal-2019-deep}.

Several researchers suggest global learning with beam-search strategy in \workcite{zhang-nivre-2011-transition} under the neural setting.
\newcite{zhou-etal-2015-neural} make the pioneer attempts for this goal,
which is further perfected with a theoretical guaranty by \newcite{andor-etal-2016-globally}.
These models have achieved state-of-the-art performance before the biaffine parser \cite{dozat2016deep}.
One major drawback is that the strategy suffers from the efficiency problem due to the beam search.
The dynamic oracle strategy is applied as well making the greedy transition-based neural dependency parsers
\cite{fernandez-gonzalez-gomez-rodriguez-2018-dynamic-oracle}.
Recently, both global learning and dynamic oracle are difficult to give much-improved capacity
when pretrained contextualized word representations are exploited.

\subsection{Other Frameworks}
Several interesting models outside the graph-based and transition-based framework are also concerned.
For example, the grammar-based framework can be applied to dependency parsing as well.
First, a dependency tree is converted to an equivalent phrase-structural constituent tree,
and then a grammar-based constituent parsing model can be applied for dependency parsing.
The method is proposed firstly by \newcite{mcdonald-06-phd-thesis},
and also highly emphasized in \newcite{kubler2009dependency}.
Several studies have exploited this method as one component for model ensembling \cite{sun-wan-2013-data}.
Recently, \newcite{zhou-zhao-2019-head} and \newcite{mrini2019rethinking} adopt the HPSG grammar for the same goal,
achieving very competitive performances.

\newcite{goldberg-elhadad-2010-efficient} present an easy-first dependency parsing model, which processes the input sentences in a non-directional way.
The output dependency tree is constructed recursively, where the highest-confidence dependency arc is selected at each time.
The neural version of the work is exploited by \newcite{kiperwasser-goldberg-2016-easy} by using hierarchical LSTMs.
Sequence-to-sequence learning can be also applied to neural dependency parsing, where the transition-based linearization can be served as one natural solution.
\newcite{li-etal-2018-seq2seq} present a strong sequence-to-sequence model by head prediction for each word.
\newcite{strzyz-etal-2019-viable} suggest a sequence labeling model for dependency parsing.

\subsection{Semi-Supervised Models}
Here we briefly offer a survey for semi-supervised dependency parsing under the traditional statistical setting,
which utilizes statistical information extracted from a raw text to enhance a baseline model.
This scheme has received little attention under the neural setting because of pretraining.
As a whole, the semi-supervised dependency parsing models can be categorized into three types
according to the extracted information from the raw text,
namely word-level, partial-tree level, and sentence-level methods.

For word-level information, one representative work is \workcite{koo-etal-2008-simple},
which augments the atomic features of a baseline model with word clusters.
\newcite{zhou-etal-2011-exploiting} exploit selectional preference information from web texts to improve dependency parsing.
Actually, word embeddings can be also regarded as a kind of semi-supervised word-level information,
which has been suggested by \newcite{turian-etal-2010-word} for NLP, but not experimented on dependency parsing.
\newcite{chen-etal-2014-feature} further extend the idea into feature embeddings, embedding all features including words.

For the partial-tree level integration, \newcite{chen2008dependency}
exploit high-frequency auto-parsed bilexical dependencies to enhance the baseline supervised model.
Further, \newcite{chen-etal-2009-improving} extend the work by using higher-order subtrees.
\newcite{chen-etal-2012-utilizing} could be regarded as a general framework for the partial tree level integration,
by utilizing dependency language models learned from auto-parsed dependency trees.

Self-training, co-training as well as tri-training are straightforward methods for sentence-level
semi-supervised learning \cite{sogaard-rishoj-2010-semi},
where high-confidence auto-parsed dependency trees from several baseline models,
are used to augment the training dataset.
\newcite{li-etal-2014-ambiguity} propose an ambiguity-aware learning method
to effectively model the confidence of auto-parsed dependency trees,
leading to significant performance improvements.

\subsection{Model Ensemble}
By effectively combining heterogeneous models,
the dependency parsing performance can be further boosted.
\newcite{nivre-mcdonald-2008-integrating} first analyze the differences between graph-based and transition-based models,
and then combine the two kinds of models to utilize their complementary information,
resulting in better performances.
\newcite{sun-wan-2013-data} perform parsing ensemble by including grammar-based models further,
which are highly diverse with the graph-based and transition-based models.
Under the neural setting, simple voting can achieve very strong performances.

The above studies are all targeted at different parsing models based on the same treebank.
There are several studies aimed at the parser ensemble based on heterogeneous treebanks,
whose annotation guidelines are highly different.
\newcite{li-etal-2012-exploiting} exploit stacked learning combine with quasi-synchronous grammars for effective integration.
\newcite{guo-etal-2016-universal} study a similar ensemble by using deep multitask learning,
where treebanks of different languages are also concerned.
\newcite{jiang-etal-2018-supervised} present and study the task of supervised treebank conversion,
which can be served as one method for integration.

\section{Semantic Dependency Graph}
The dependency parsing models mentioned in the previous section are all aimed for dependency tree parsing,
which majorly reflects syntactic and shallow-semantic information in sentences.
As there are growing demands of deep semantic parsing,
which is difficult to be expressed by dependency tree only,
dependency graph parsing has received increasing interests \cite{oepen-etal-2014-semeval,oepen2015semeval,che-etal-2016-semeval},
which allows multiple (including zero) heads for one word in sentences.
Note that the semantic graph is still formalized by a set of bilexicalized dependencies,
with nodes corresponding to surface lexical words, and edges indicating the semantic relations between nodes.

There are different formalizations of the semantic dependency graph.
We can combine syntactic tree-based dependency parsing and semantic role labeling (SRL) to result in a dependency graph,
which is referred to as joint dependency syntax and SRL \cite{surdeanu-EtAl:2008:CONLL,hajic-etal-2009-conll1}.
Recently, the conception of semantic dependency parsing (SDP) has been introduced \cite{oepen-etal-2014-semeval,oepen2015semeval,che-etal-2016-semeval},
which provides different views of semantic relations, such as DELPH-IN MRS (DM), 
predicate-argument structures (PDS)  and Prague semantic dependencies (PSD).
Following, we will review the studies of the two types of semantic dependency graph parsing.

\begin{figure}[H]
\includegraphics[scale=0.7]{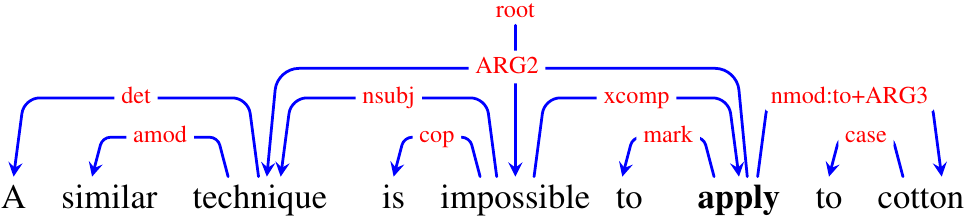}
\caption{An example of joint syntactic and semantic dependencies.}
\label{fig-depsrl}
\end{figure}

\subsection{Joint Dependency Syntax and SRL} \label{sec-dep-srl}
Figure \ref{fig-depsrl} shows an example dependency graph of joint syntactic and semantic dependencies.
Here we do not intend to introduce the pipeline models,
which train syntactic and semantic models separately,
and then output the dependency graph by either two steps or jointly \cite{che-EtAl:2009:CoNLL-2009-ST,johansson-2009-statistical}.
Although these models can perform dependency graph parsing,
they receive less attention as this topic.
We focus on the models of joint learning and decoding for full dependency graph parsing.
Table \ref{table:jointsrl:performance} shows the performance of several studies on this line.

\setlength{\tabcolsep}{2.2pt}
\begin{table}[H]
\begin{threeparttable}
\footnotesize
\caption{A comparison of typical joint dependency syntax and SRL models on the CONLL08 English dataset.  } \label{table:jointsrl:performance}
\begin{tabular}{l|c|ccc}
\hline
Model &  Main Features &  Syn &  Sem  & All \\ \hline \hline
\sworkcite{johansson-2009-statistical} &  joint inference & 86.6 & 77.1 & 81.8 \\
\sworkcite{titov2009online} &  transition-based &  87.5  & 76.1 & 81.8\\
\sworkcite{henderson-etal-2013-multilingual}  &  sigmoid belief network  & 87.5  & 76.1 & 81.8 \\
\sworkcite{swayamdipta-etal-2016-greedy} & neural, stack-LSTM  & \bf 89.1  & \bf 80.5  & \bf 84.5\\
\hline
\end{tabular}
\end{threeparttable}
\end{table}

\newcite{titov2009online} extend the transition-based dependency parsing with a particular \emph{swap} operation,
enable the model to process non-planarity multiple graphs jointly,
and thus dependency graph parsing can be performed jointly.
\newcite{henderson-etal-2013-multilingual} also exploit the transition-based framework to derive syntactic and semantic
dependencies concurrently based on a similar transition system as \newcite{titov2009online},
but adopt a different model estimation by using an incremental sigmoid belief network with latent variables.
\newcite{lluis-etal-2013-joint} present a graph-based model with a dual decomposition algorithm for decoding,
assigning syntactic and semantic dependencies concurrently.

All the aforementioned studies are based on the traditional statistical setting.
Under the neural setting, there is little work focus on the task, with one exception.
\newcite{swayamdipta-etal-2016-greedy} present a transition-based stack-LSTM model for joint syntactic and semantic dependencies,
where their transition system is largely followed \workcite{henderson-etal-2013-multilingual}.
Since then, neural dependency graph dependency parsing models are centered on other datasets.

\begin{figure}[H]
\includegraphics[scale=0.75]{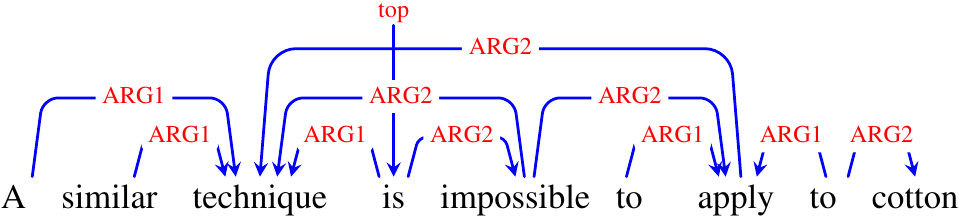}
\caption{An example of semantic dependency graph.}
\label{fig-depgraph}
\end{figure}

\setlength{\tabcolsep}{3.0pt}
\begin{table}[H]
\begin{threeparttable}
\footnotesize
\caption{A comparison of typical dependency parsing models on the SemEval-2015 shared dataset, where WSJ and Brown indicate the in-domain and out-of-domain test sections. } \label{table:sdp:performance}
\begin{tabular}{l|c|cc}
\hline
Model &  Main Features &  WSJ &  Brown   \\ \hline \hline
\sworkcite{du2015peking} &  tree approximations & 85.4 & 80.8  \\
\sworkcite{almeida2015lisbon} &  2-order graph &  85.2  & 81.2 \\
\sworkcite{peng-etal-2017-deep}  &  multi-task learning  & 87.2  & 83.6  \\
\sworkcite{wang2018neural} & transition, LSTM  & 86.9  &  82.8  \\
\sworkcite{dozat-manning-2018-simpler} & LSTM, biaffine  &  89.5  &  86.3  \\
\sworkcite{wang-etal-2019-second} & 2-order graph, LSTM  & \bf 89.8  & \bf 86.9  \\
\hline
\end{tabular}
\end{threeparttable}
\end{table}

\subsection{Semantic Dependency Parsing}
SDP could be regarded as an extension from syntactic dependency parsing
by characterizing more semantic relations over the bilexical dependencies \cite{sun-etal-2014-grammatical,che-etal-2016-semeval},
which can be greatly benefited from the advances of dependency parsing.
While recently, \newcite{oepen-etal-2014-semeval} and \newcite{oepen2015semeval} present SDP from a different view,
which converts the already available linguistic-informed semantic annotations into dependencies,
including three different formalisms: DM, PAS and PAD,
and currently it has been widely accepted for deep semantic parsing.
Figure \ref{fig-depgraph} shows an example of SDP.
For SDP, graph- and transition-based models are also the mainstream methods,
and most of these models are adapted from dependency tree parsing.
Table \ref{table:sdp:performance} shows the performance of several representative SDP models.

\subsubsection{Graph-based}
There are a range of graph-based SDP models for the shared tasks of SDP in SemEval \cite{thomson-etal-2014-cmu,almeida2015lisbon}.
Generally, it is hard to develop a graph-based decoding algorithm targeted to arbitrary dependency graphs.
Thus, most models have imposed particular constraints.
\newcite{kuhlmann-jonsson-2015-parsing} present a cubic-time exact inference algorithm for non-crossing dependency graphs.
\newcite{cao-etal-2017-parsing} and \newcite{cao-etal-2017-quasi} investigate the maximum subgraph algorithm for 1-endpoint-crossing, pagenumber-2 graphs.
\newcite{sun-etal-2017-parsing} attempt to solve the dependency graph parsing by subgraph decomposition and merging.
\newcite{sun-etal-2017-semantic} propose an interesting book embedding strategy for SDP.

All the above models exploit manually-crafted discrete features.
Under the neural setting, \newcite{peng-etal-2017-deep} present a multi-task learning framework to different views of SDP.
\newcite{dozat-manning-2018-simpler} extend the biaffine dependency parsing for SDP.
Recently, \newcite{wang-etal-2019-second} propose a second-order SDP model based on \workcite{dozat-manning-2018-simpler}.
As a whole, neural models can obtain better performances for SDP.

\subsubsection{Transition-based}
The transition-based SDP models can also achieve competitive performance, and meanwhile, these models are more efficient and free of constraints,
thus they have received great attention \cite{ribeyre-etal-2014-alpage,kanerva-etal-2015-turku}.
Actually, transition-based dependency graph parsing can be dated back to \newcite{sagae-tsujii-2008-shift},
and the model is enhanced with dynamic oracle by \newcite{tokgoz2015transition}.
\newcite{sun-etal-2014-grammatical} define a K-permutation transition system to handle dependency graph generation.
\newcite{zhang2016transition} present two novel transition systems for deep semantic dependency parsing.
\newcite{gildea-etal-2018-cache} presents a transition-based system by including a cache to capture dependency graphs,

Recently, \newcite{wang2018neural} propose a strong transition-based SDP model by using neural networks.
They exploit deep bidirectional LSTM as sentential encoder together with stack-LSTM for better representation of transition states.
\newcite{buys-blunsom-2017-robust} present a transition-based model for general semantic graph parsing,
which is also suitable for SDP.

\subsubsection{Other Methods}
Dependency graph parsing by using tree approximations and post-processing is also able to obtain competitive performance.
These kinds of models first convert dependency graphs into trees, and then tree-based parsing can be applied \cite{agic-koller-2014-potsdam,schluter-etal-2014-copenhagen}.
\newcite{du2015peking} ensemble several tree approximation strategies and achieve the top performance in SemEval 2015 \cite{oepen2015semeval}.
\newcite{agic2015semantic} conduct a comprehensive investigation of semantic dependency graph parsing using tree approximations.

\begin{figure}[H]
\includegraphics[scale=0.65]{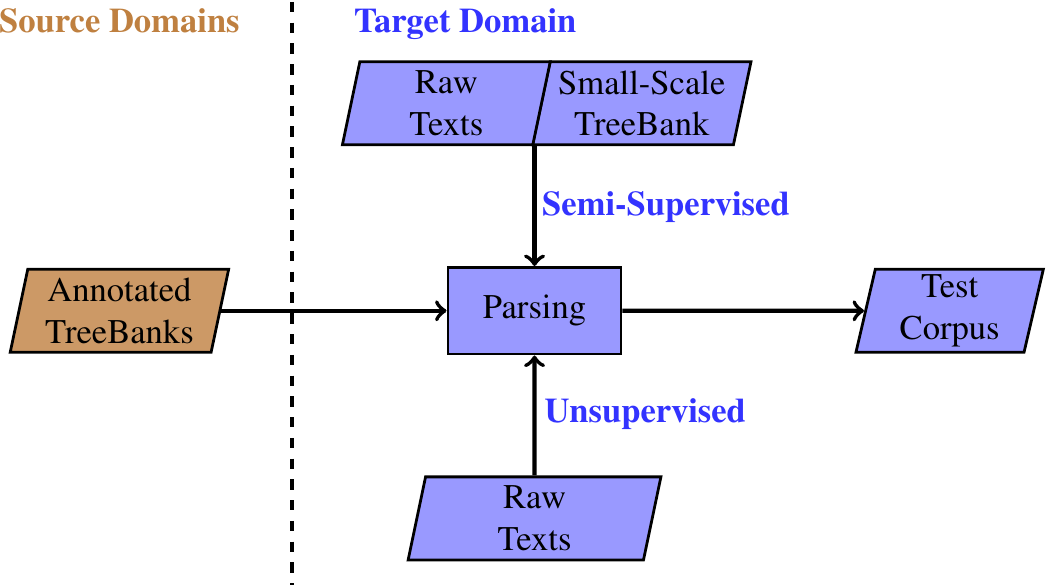}
\caption{The architecture of cross-domain parsing.}
\label{fig-cross-domain}
\end{figure}

\section{Cross-Domain Parsing}
Cross-domain adaption is one hot topic in the NLP community,
especially for the syntactic and semantic parsing tasks,
where the data annotation is extremely laborious and expensive.
Currently supervised parsing has achieved incredibly high performances thanks to the recent advances of neural networks.
However, the performance could drop significantly when the well-trained parsers are applied to texts in different domains as the training corpus.
It is impractical to annotate training datasets for all domains.
Thus, cross-domain adaption is very important to make parser applicable.
The studies of cross-domain parsing are focused on two settings majorly: unsupervised domain adaption,
where no target domain training dataset is available,
and semi-supervised domain adaption,  where a small-scale of training instances are available for a target domain.
Figure \ref{fig-cross-domain} shows the architecture of cross-domain parsing, where the differences between the two settings are illustrated.

\subsection{Unsupervised Domain Adaption}
Self-training is one useful strategy for cross-domain parser adaption,
although it has achieved very limited gains under the in-domain semi-supervised setting.
Initial work mostly focuses on constituent parsing.
\newcite{mcclosky-etal-2006-reranking} exploit a reranking strategy to
obtain a set of high-confidence auto-parsed outputs, and then add them to the training corpus.
\newcite{sagae-2010-self} shows that without reranking self-training alone can also give significant improvements.
\newcite{kawahara-uchimoto-2008-learning} firstly apply self-training successfully on dependency parsing,
which exploits an extra classifier to determine whether a parsed tree is reliable.
\newcite{chen-etal-2008-learning} exploit only high-confidence partial dependencies for next-round training.
\newcite{yu2015domain} propose a novel confidence estimation method,
leading to improved performance on the out-of-domain dataset.

Besides self-training, there are several other methods for unsupervised domain adaption.
\newcite{steedman-etal-2003-example} apply co-training to constituent parsing, which is similar to self-training
but difference in that the example selection is performed by two parsers.
\newcite{sagae-tsujii-2007-dependency} use a similar co-training method for dependency parsing.
Further, \newcite{sogaard-rishoj-2010-semi} exploit tri-training for domain adaption of dependency parsing,
extending two parsers into parsers.
Interestingly, \newcite{plank-van-noord-2011-effective} select training instances
from the source-domain dataset instead, where the instances most relevant to the target domain are chosen.
\newcite{yang-etal-2015-domain} exploit deep belief neural networks to enhance the dependency parsing performance on out-of-domain test data,
which can effectively extract useful information from target-domain raw texts.

Multi-source domain adaption is also a promising direction,
which assumes that training corpora of several source domains are available.
The setting is highly matched with the real practical scenario.
\newcite{mcclosky-etal-2010-automatic} present the first work of this setting for dependency parsing.
They linearly combine the parsing models of different domains
with the weights learned from a regression model,
considering the performance of each individual parser on the target domain.

\subsection{Semi-Supervised Domain Adaption}
With a small number of target domain training dataset,
\newcite{reichart-rappoport-2007-self} show that self-training can effectively improve the performance of constituent parsing.
Recently, most work focuses on effectively training on the mixed source and target training instances
by separating the domain-dependent and domain-invariant features \cite{daume-iii-2007-frustratingly}.
By treating these features differently, the final model can accurately transfer the useful knowledge
from the source domain into the target.
\newcite{finkel-manning-2009-hierarchical} extend the idea with a hierarchical Bayesian model
and evaluate it on dependency parsing, achieving better performance on the target domain
than training with only the target-domain data.
Under the neural setting, adversarial learning is one effective method for the same purpose \cite{ganin2015unsupervised}.
\newcite{sano2017adversarial} firstly apply the method on dependency parsing.

Active learning can be one promising approach for semi-supervised domain adaption.
Considering that full-sentence syntax/semantic annotation is extremely expensive,
partial annotation might be preferable.
For constituent parsing, \newcite{joshi-etal-2018-extending} suggest partial annotation
of constituent brackets to enhance domain adaption.
For dependency parsing, \newcite{flannery2015combining} exploit partial annotation
combined with active learning  for cross-domain dependency parsing in Japanese.
Recently, \newcite{li-etal-2019-semi-supervised} investigate the strategy comprehensively for Chinese dependency parsing
under the neural setting.

\begin{figure}[H]
\includegraphics[scale=0.65]{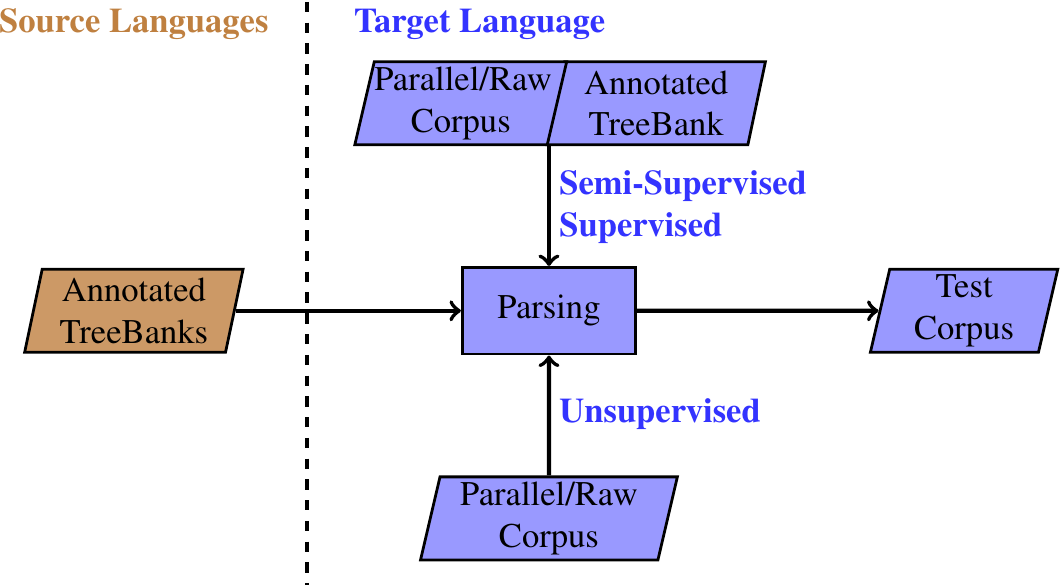}
\caption{The architecture of cross-lingual parsing.}
\label{fig-cross-lingual}
\end{figure}

\section{Cross-Lingual Parsing}
Cross-lingual parsing, which aims to parse the sentence structures of low-resource languages
with the help of resource-rich languages such as English.
There have been a number of studies for this task,
and the majority of work focuses on dependency parsing
due to the relatively structural conciseness as well as well-developed universal dependencies.
In particular, with the recent development of cross-lingual or universal
word representations based on neural pretraining techniques,
the task has been concerned with increasing interests.
The task includes two main settings,
the unsupervised setting assuming that no training corpus is available for target languages,
and the semi-supervised/supervised setting where there exists a certain scale of corpora for the target languages.
The architecture of cross-lingual parsing is shown in Figure \ref{fig-cross-lingual},
where the detailed difference between unsupervised and semi-supervised/supervised settings are illustrated as well.

\subsection{Unsupervised Setting}
For unsupervised cross-lingual parsing,
the mainstream methods can be classified into two categories,
model transferring and annotation projection,
where the first category trains a model on the source-language training corpus,
and then directly uses it to parse the target-language texts,
and the second category
projects the source-language parse annotations into the target-language by using a parallel corpus,
resulting in a pseudo training corpus for the target language,
and then trains a target-language parsing model on the pseudo corpus.

\subsubsection{Model Transferring}
The model transferring approach is straightforward for cross-lingual parsing.
The most effort is concerned with language-independent features,
which play consistent functions across languages.
This line of work is initially presented by \newcite{zeman-resnik-2008-cross} which suggests delexiciallized models for cross-lingual dependency parsing,
and is further developed by \newcite{mcdonald-etal-2011-multi} for multi-source transferring,
where multiple source languages are used to enhance a target language.
Several researchers resort to various non-lexical features to enhance the delexicalized cross-lingual models
\cite{cohen-etal-2011-unsupervised,naseem-etal-2012-selective}.

Recently, \newcite{tackstrom-etal-2012-cross} exploit cross-lingual word clusters,
which is one king of cross-lingual word representations.
Under the neural setting, the exploration of cross-lingual word representations is greatly facilitated.
\newcite{guo-etal-2015-cross} propose to use cross-lingual word embeddings for lexicalized cross-lingual dependency parsing.
This method is then received much attention and
can be further enhanced by various ways such as better neural structures \cite{zhang-barzilay-2015-hierarchical}
and multi-source adaption \cite{guo2016representation,wick2016minimally}.

Cross-lingual pretrained contextualized word representations give the state-of-the-art performances of this category.
\newcite{schuster-etal-2019-cross} provide a method to learn contextual ELMO representations effectively and
then apply the representations on the task, achieving much better performances than cross-lingual word embeddings.
\newcite{wang-etal-2019-cross} and \newcite{wu-dredze-2019-beto} apply cross-lingual mBERT
to zero-shot cross-lingual dependency parsing.
\newcite{lample2019cross} introduce the XLM concurrently to mBERT, which is also a kind of
strong multilingual contextualized word representations for cross-lingual parsing \cite{wu2019emerging}.
All these recent studies lead to state-of-the-art performances in the literature of this category.

\subsubsection{Annotation Projection}
The annotation projection approach requires slightly more effort compared with model transferring,
which aims to build a pseudo training corpus through bitext projection.
With the pseudo training corpus, the final model can capture rich target-language characteristics.
The method relies on a set of parallel sentences between the source and target languages.
A source parser trained on the source treebank is used to parse the source-side sentences of the parallel corpus,
and then the automatic source annotations are projected onto the target language sentences according to word alignments,
resulting in the final pseudo training corpus.
There are a range of strategies to achieve the goal.
For example, we can use different kinds of parallel corpora,
such as EuroParl and the book Bible,
and can also exploit various sophisticated methods to improve the projection quality.

For constituent parsing,  \newcite{snyder-etal-2009-unsupervised} exploit the method for unsupervised constituent parsing,
and find that it can significantly outperform the purely-unsupervised models.
\newcite{jiang-etal-2011-relaxed} suggest an EM algorithm to incremental boost the quality of the projected constituent trees with relaxing constraints.
The number of studies on constituent parsing is relatively small,
which may be possible due to that the projection of constituent structures is very complex.

For dependency parsing, \newcite{hwa2005bootstrapping} present the first work of this category,
and then the approach has been extensively studied under different settings, such as
confidence-aware learning \cite{li-etal-2014-soft},
neural network enhancing \cite{ma-xia-2014-unsupervised,schlichtkrull-sogaard-2017-cross},
and multi-source adaption \cite{rasooli-collins-2015-density,agic-etal-2016-multilingual}.
Interestingly, \newcite{jiang2015joint} propose a joint model for cross-lingual constituent and dependency parsing with annotation projection.
The approach achieves great success for cross-lingual dependency parsing.

\subsubsection{Other Methods}
There are also several other methods for unsupervised cross-lingual parsing.
Treebank translation is one representative strategy,
which is essentially highly similar to annotation projection.
The approach also aims to construct a pseudo training corpus.
Different from annotation projection, it directly translates the source training corpus into the target language.
Besides bitext projection, it requires translation to produce target language sentences.
\newcite{tiedemann2014treebank} firstly propose this method
and their method is further perfected by their later studies \cite{tiedemann2016synthetic}.
\newcite{zhang-etal-2019-cross} study the approach under the neural setting with partial translation,
and combine their model with model transferring.

The methods exploited in cross-domain parsing may be also suitable (e.g., self-training) for this setting
because of the cross-lingual word representations.
However, these kinds of methods have been seldom studied.
\newcite{rasooli-collins-2017-cross} combine the advantages of model transferring, annotation projection,
treebank translation as well as self-training to obtain a very strong model for cross-lingual dependency parsing.

Sentence reordering is one interesting method presented recently, which aims to
reorder the input source language syntactic trees to make it highly similar to the target language.
The idea is first studied by \newcite{wang-eisner-2018-synthetic}.
\newcite{rasooli-collins-2019-low} exploit the method with two strong reordering strategies,
obtaining very competitive performance compared with even supervised parsing models.

\subsection{Semi-Supervised/Supervised Setting}
As the availability of treebanks for a range of languages,
how to effectively exploit both source and target language treebanks
is one interesting problem.
Since very early, several studies show that two languages are better than one language alone for parsing.
\newcite{smith2004bilingual} show that joint training the English and the Korean parser can bring better performance.
\newcite{burkett-klein-2008-two} also demonstrate the same observation.

Under the neural setting, this line of work can be conducted more conveniently due to the cross-lingual word representations.
\newcite{ammar-etal-2016-many} propose to use one single universal model to parse all languages.
However, their final performance is still below the corresponding individual baselines.
\newcite{smith-etal-2018-82} train 34 models for 46 different languages.
By aggregating multiple treebanks from one language or closely related languages,
we can achieve competitive performances and meanwhile reduce the number of required models greatly.
Most recently, \newcite{kondratyuk-straka-2019-75} propose a sophisticated strategy to train one universal model for 75 languages by leveraging a multilingual BERT self-attention,
which achieves better performances than the corresponding individual models.

\section{Joint Models}
\begin{figure}[H]
\includegraphics[scale=0.75]{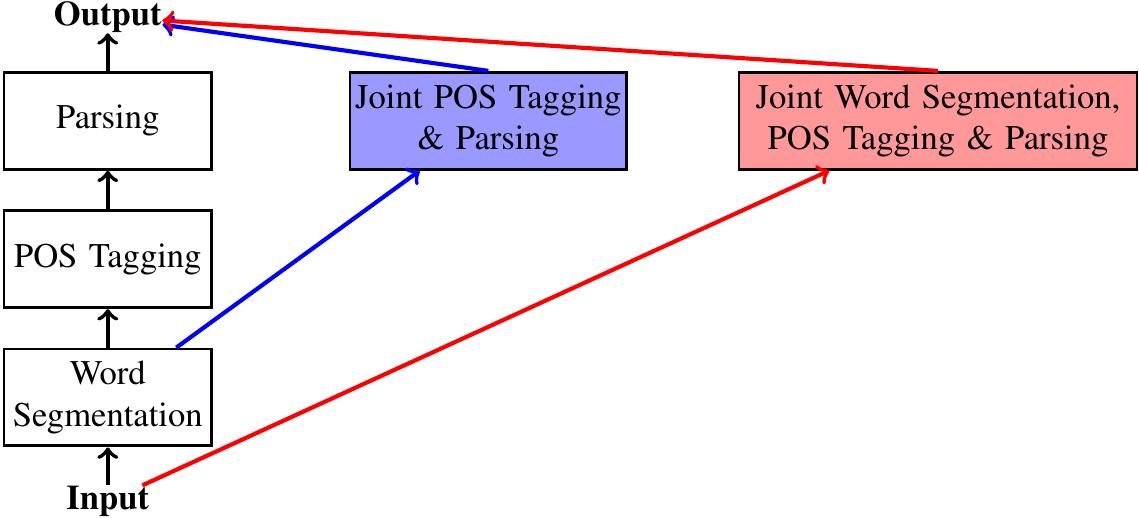}
\caption{The architecture of joint models targeted for parsing, where word segmentation is only available to the Chinese language.}
\label{fig-joint-model}
\end{figure}

In this section we discuss joint models of parsing,
focusing only on the final goal being the parsing task.
The studies of jointly modeling syntax-semantic parsing as well as a targeted downstream task will be introduced in the next section.
The development of joint models is mainly motivated by the error prorogation problem of the preconditioned tasks.
POS tagging is one of the major preconditioned tasks,
as POS tags are one kind of valuable feature source for parsing.
Before POS tagging, several languages such as Chinese require word segmentation as a prerequisite step.
Parsing is generally performed based on words, while sentences of these languages do not have explicit word boundaries.
In summary, here we briefly investigate two kinds of joint models: joint POS tagging and parsing,
joint segmentation \& tagging and parsing,
and we show their relationship in Figure \ref{fig-joint-model}.

Noticeably, there are several studies for joint syntactic and semantic parsing.
The dependency-based joint models have been already described in Section \ref{sec-dep-srl}.
Thus, one can refer to there for details.
For joint constituent parsing and semantic role labeling,
there are very few studies.
The representative work is \newcite{li-etal-2010-joint},
which is the first work of this kind by using sophisticated manually-crafted features.
The work shows that their joint model is able to give better performances for both Chinese constituent parsing and SRL.

\subsection{Joint POS Tagging and Parsing}
For joint POS tagging and constituent parsing,
the chart-based PCFG parsing naturally performs the two tasks concurrently \cite{collins-1997-three,Charniak2000,petrov-klein-2007-improved},
where POS tags can be directly induced from the bottom lexical rules.
Based on the transition-based framework, joint POS tagging and constituent parsing can be
easily achieved by the shift operation with one additional parameter to indicate the POS tag of the processing word.
\newcite{wang-xue-2014-joint} investigate the joint task and present a number of non-local features.

\newcite{li-etal-2011-joint} propose the first joint model of POS tagging and dependency parsing based on graph factoring,
where the basic scoring units are augmented with POS tags.
\newcite{li-etal-2012-separately} enhance the model with better learning strategies.
\newcite{hatori-etal-2011-incremental} is the first transition-based model for joint POS tagging and dependency parsing.
\newcite{bohnet-nivre-2012-transition} extend the transition-based model for non-projective dependency parsing.
The two kinds of models achieve comparable performances for both tasks.

Under the neural setting, \newcite{alberti-etal-2015-improved} investigate
the model of \newcite{bohnet-nivre-2012-transition} with neural features.
\newcite{zhang-weiss-2016-stack} suggest a joint POS tagging and dependency parsing model by stack propagation.
\newcite{yang2017joint} further investigate the neural joint task with LSTMs
by using graph-based and transition-based frameworks, respectively.
In fact, the importance of joint modeling has been largely weakened
as parsing without POS tags can also obtain strong performance which is close to the same model with POS tags \cite{dozat2016deep}.

\subsection{Joint Segmentation, Tagging and Parsing}
The task of joint segmentation, tagging and parsing is majorly targeted to Chinese parsing.
The series of this work starts very early \cite{Luo:2003:EMNLP} by character-level parsing.
Later, \newcite{zhao-2009-character} demonstrate that Chinese dependency parsing based on characters is better,
which can naturally perform the three tasks.
Recently, \newcite{hatori-etal-2012-incremental} propose a transition-based joint model for word segmentation, POS tagging and dependency parsing.
\newcite{li-zhou-2012-unified} suggest a similar transition-based joint model by using indivisible subwords as well as their internal structures.
\newcite{zhang-etal-2013-chinese} and \newcite{zhang-etal-2014-character} conduct character-level constituent and dependency parsing by extending word-level annotations into characters,
achieving state-of-the-art performances for both tasks under the discrete setting.
All the four models exploit transition-based framework.
\newcite{zhang-etal-2015-randomized} propose the first work by using graph-based inference, with efficient hill-climb decoding.

\newcite{zheng2015character} is the first work of adopting neural networks for character-level constituent parsing,
achieving comparable with the state-of-the-art discrete model by a simple convolutional neural network.
\newcite{li2018neural} present a neural model for character-level dependency parsing.
\newcite{yan2019unified} propose a strong joint model for word segmentation and dependency parsing only,
state-of-the-art biaffine parser and pretrained BERT are exploited in this work.
Under the neural network, the joint framework might be highly challenging,
as the baselines are strong and meanwhile neural networks can learn global high-level features implicitly.

\section{Parser Application}
When a well-trained syntactic/semantic parser is available,
how to use it effectively to benefit for downstream applications is one important topic in the parsing community,
which is also related to the verification of the usefulness of syntactic and semantic parsing.
In fact, the topic has been extensively studied,
and the parsing outputs have been demonstrated effective for a number of tasks such as semantic role labeling \cite{johansson-nugues-2008-dependency,strubell-etal-2018-linguistically},
relation extraction \cite{zhang-etal-2006-exploring,miwa-bansal-2016-end}, sentiment analysis \cite{zou2015sentiment,tai-etal-2015-improved}
and machine translation \cite{yamada-knight-2001-syntax,zhang-etal-2019-syntax-enhanced-neural}.
The exploration methods have major changes from the statistical discrete models
to the neural models.
Here we briefly summarize the mainstream approaches of parser exploration in terms of the two settings.

\subsection{Feature-Based Statistical Methods}
Under the traditional statistical setting,
the exploration of parser resorts to manually-crafted discrete features,
which are mostly designed sophisticatedly according to the targeted tasks.
We briefly summarize the widely-adopted features here.
For constituent trees, such features include non-terminal categories,
CFG rules, phrase-level word ngrams, syntax paths to the root or some other word,
the matching with a completed phrase.
For dependency trees, dependency-based ngrams, dependency labels, dependency paths
are widely-used features.
All these kinds of features are further adapted to various tasks aiming
to get most of the parsing information effectively \cite{liu-etal-2006-tree,johansson-nugues-2008-dependency,chan-roth-2011-exploiting,qiu-zhang-2014-zore,zou2015sentiment}.
Besides, the tree-kernel based approach can also be good alternatives \cite{che-etal-2006-hybrid,yang-etal-2006-kernel,zhang-etal-2006-exploring,zhou-etal-2007-tree,zhang-li-2009-tree}.
Several approaches suggest using multiple heterogeneous parsers for better performances,
including the integration of constituent and dependency parsers as well as parsers trained on heterogeneous treebanks \cite{johansson-nugues-2008-effect}.

\subsection{Representation Learning with Neural Networks}
\begin{figure}[H]
\includegraphics[scale=0.75]{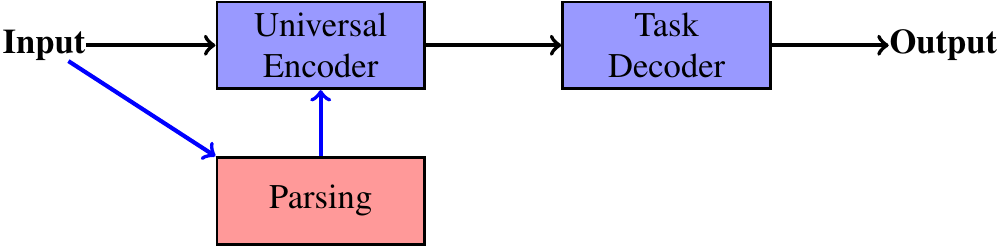}
\caption{Parser enhanced universal encoder for downstream tasks.}
\label{fig-parser-apply}
\end{figure}

One simple method to use parsing features based on neural networks is to embed all the atomic features,
and then exploit sophisticated neural networks to compose them automatically.
The most representative method of this kind is the path-based LSTMs,
which exploit LSTM over sequential-level constituent or dependency paths \cite{xu-etal-2015-classifying,roth-lapata-2016-neural}.
The recent tendency of using the end-to-end framework for the majority of NLP tasks
leads to universal representations based on parser outputs.
We build a universal encoder with structural outputs of a parser, and then adapt them to different tasks by decoders, as shown by Figure \ref{fig-parser-apply}.
There are several ways to build the encoder.
Here we divide the methods into four types: recursive neural network; linearization-based; implicated structural-aware word representations and graph neural networks (GNN).

The recursive neural network is one natural method to model tree-structural outputs,
which composes a tree input from bottom-to-up or top-to-down incrementally.
We can use various composition operations leading to more sophisticated tree-level neural networks
such as tree convolutions suggested by \newcite{mou-etal-2015-discriminative} and Tree-LSTM proposed by \newcite{tai-etal-2015-improved}.
All these related studies give improved performances for a range of tasks \cite{zhang-etal-2016-top,teng-zhang-2017-head}.

The key idea of the linearization-based methods is to convert structural inputs into a sequence of symbols,
and then adopt standard sequential encoders to model the new sequence directly \cite{li-etal-2017-modeling,wu2017improved}.
Usually, the conversion can be referred to as the linearization process of transition-based parsers,
or we can incrementally traverse a tree or graph in different ways.
The method has received fewer concerns which might be due to its extreme simplicity,
although it is effective and meanwhile much efficient \cite{zhang-etal-2019-syntax-enhanced-neural}.

The implicit structural-aware word representations, firstly presented by \newcite{zhang-etal-2017-end} for relation extraction, are similar to the idea of contextualized word representations,
which exploit the hidden outputs of a well-pretrained parser as inputs for the downstream tasks \cite{yu-etal-2018-transition,zhang-etal-2019-syntax-enhanced-neural}.
This method can also efficiently represent structural information such as syntax and semantics.
Besides, the method can be easily adapted to the multi-task-learning strategy for
parser application \cite{strubell-etal-2018-linguistically},
while parser requires to be jointly trained in multi-task-learning.

Recently, there are grown interests on the topic of graph neural networks,
which can be naturally applied to encode structural syntactic and semantic graphs.
Indeed, there have been several studies already
by using either graph convolutional networks or graph attention networks \cite{bastings-etal-2017-graph,zhang-etal-2018-graph,marcheggiani-etal-2018-exploiting},
and all these works demonstrate the effectiveness of GNN for structure encoding.

\section{Corpus and Shared Tasks}
Finally, we review the work of corpus development in syntactic and semantic parsing,
which is critical to the performance of supervised parsing.
There are several classical treebanks such as the Penn Treebanks of English and Chinese languages,
which greatly promote the development of the parsing community.
In fact, there are treebanks for a range of languages,
and here we focus majorly on the Chinese and English treebanks.
In addition, there are a number of shared tasks,
which also offer valuable corpora for syntactic and semantic parsing.

\subsection{Penn Treebank}
The English Penn Treebank (PTB) by \newcite{marcus-etal-1993-building} could be the most famous resource for syntactic parsing,
which annotates bracketed syntactic phrase structures for over 40,000 sentences covering about 4.5 million words.
In addition, \newcite{xuexia2005} annotate the Penn Treebank for the Chinese language, for short as CTB,
and now there are over 130,000 sentences with phrase-structure annotations covering over 2 million words.
Both the two datasets have annotated POS tags as well, which are important to automatic syntactic parsing.
For Chinese, gold-standard word segmentation has been annotated in CTB as well.

The two datasets are also converted into dependency treebanks for dependency parsing,
which could be achieved by rule-based head lexicalization over the constituent trees \cite{Yamada2003,johansson2007extended,johansson2007lth,zhli2018convert,zhang-clark-2008-tale}.
Recently, Stanford dependencies have been exploited the most popularly especially for the English language,
where the conversion rules are relatively more fine-grained \cite{de2006generating} and meanwhile can reflect
more syntactic and semantic phenomena.

There are several small-scale treebanks with the same annotation guideline as PTB,
which can be useful resources for domain adaption studies of constituent and dependency parsing,
regarding that PTB are focused on the news genre data.
For example, the Brown Treebank is exploited most frequently for cross-domain parsing as the literature genre.
\newcite{tateisi2005syntax} offer a treebank of the biomedical domain.
The two treebanks are targeted to researches on constituent parsing.
Recently, \newcite{kong-etal-2014-dependency} annotate a treebank for twitter texts based on dependency grammar.

\subsection{Universal Dependencies}
The present of Universal Dependencies (UD) has received great attention for facilitating multilingual researches,
which aims to develop cross-linguistically consistent treebank annotation for multiple languages.
UD can capture similarities as well as idiosyncrasies among typologically different languages such as English-alike languages, morphologically-rich languages
and pro-drop languages.
The development of UD is initially based on Stanford typed dependencies \cite{de2014universal} and the universal Google dependency scheme \cite{petrov2012universal,mcdonald-etal-2013-universal1}.
Now it goes through several versions \cite{nivre2015universal,nivre2017universal}, with significant changes on the guidelines,
also supporting language-specific extensions when necessary.
Currently the UD treebank version 2.5 includes 157 treebanks over 90 languages.
Besides multilingual dependency parsing, there is an increasing tendency to exploit them for evaluating monolingual dependency parsing based on the datasets as well \cite{ji-etal-2019-graph,kulmizev-etal-2019-deep}.

\subsection{Chinese Treebank}
For the Chinese languages, treebank development has been concerned by several studies besides the CTB.
The Sinica Treebank has offered phrase-structural syntactic trees over about 360,000 words in traditional Chinese \cite{chen2003sinica}.
\newcite{qiang2004annotation} release a constituent treebank covering about one million words for simplified Chinese.
\newcite{zhan2012application} also annotate constituent trees over a scale of 0.9 million words for Chinese.
The guidelines of all these phrase-structural treebanks are quite different.

There are several treebank resources directly based on the dependency structure, as it is believed that dependency grammar is simpler and easier to be developed.
\newcite{liu2006building} and \newcite{che2012chinese} construct a Chinese dependency treebank consuming over 1.1 million words.
\newcite{qiu-etal-2014-multi} create a multi-view Chinese dependency treebank containing 14,463 sentences,
which is further augmented with predicate-argument information by \newcite{qiu2016dependency} for a semantic-oriented dependency treebank.
Most recently, \newcite{li-etal-2019-semi-supervised} release a large scale Chinse dependency treebank covering about 3 million words as well as different domains,
including news, web blogs,  literature texts.

\subsection{Shared Tasks}
Nearly all the shared tasks are focused on dependency parsing,
and most of which devote to multilingual parsing with the support of several treebanks in different languages.
These shared tasks, on the one hand, can evaluate the current state-of-the-art parsing models,
and on the other hand offer valuable datasets for parsing,
facilitating the future research work.

The ConLL06 organizes the first shared task for dependency parsing involving 13 languages \cite{buchholz-marsi:2006:CoNLL-X},
and domain adaption is considered later in ConLL07 \cite{nivre-etal-2007-conll}.
At ConLL08 and ConLL09, semantic dependencies extracted from SRL are integrated, leading to joint syntactic-semantic parsing \cite{surdeanu-EtAl:2008:CONLL,hajic-etal-2009-conll1}.
Recently, the shared task on ConLL 2017 starts to adopt universal dependencies for dependency parsing \cite{zeman-etal-2017-conll1},
and at ConLL 2018, 82 UD treebanks in 57 languages are included for evaluation \cite{zeman-etal-2018-conll}.
Besides ConLL, SANCL 2012 organizes a shared task on parsing English web text \cite{petrov-mcdonald:2012:SANCL},
which offers a benchmark dataset for cross-domain dependency parsing in English.
In addition, the NLPCC 2019 shared task on cross-domain dependency parsing also offers a valuable dataset in Chinese \cite{peng2019overview}.

The above shared tasks focus on syntactic dependency parsing.
For semantic dependency parsing, \newcite{che-EtAl:2012:STARSEM-SEMEVAL} present the first shared task to Chinese texts in SemEval,
where dependency trees are used in the evaluation.
\newcite{che-etal-2016-semeval} start to use dependency graphs for formal semantic representation.
For the English language, \newcite{oepen-etal-2014-semeval} organize a shared task for broad coverage semantic parsing by using three distinct dependency-based semantic formalizations.
Dependency graphs are exploited to represent various semantics.
\newcite{oepen2015semeval} extend the shared task of \workcite{oepen-etal-2014-semeval} with more languages including Chinese and Czech.
\newcite{oepen-etal-2019-mrp} cover more topics of semantic graph parsing for deep semantics, including not only dependency-based graphs,
but also several other formalizations such as UCCA and AMR.

\section{Conclusion and Future Directions}

In this article, we made a thorough review of the past work of syntactic and semantic parsing focusing on constituent parsing and dependency parsing.
Traditional statistical models, as well as currently-dominant neural network methods, were both summarized.
First, for the parsing models, neural network methods with pretrained contextualized word representations have achieved the top performances
for almost all datasets.
There is a grown tendency to use simple encoder-decoder frameworks for parsing,
so that well-investigated training strategies can be applied.
Second, broad-coverage semantic parsing is receiving increasing attention, which might be the next stage hop topic.
The task performances are now gradually acceptable
thanks to the neural network models as well as the development of linguistic resources.

The cross-domain and cross-lingual settings are important scenarios for parsing,
which are difficult to be resolved yet play the key role to the real applications.
For the cross-domain setting, there is still a large demand for resources.
While for cross-lingual parsing, there exist a number of methods.
A comprehensive and fair comparison of these methods as well as their integrations might be valuable.
In addition, the difference between cross-domain and cross-language is becoming smaller
because of the universal word representations.
One can regard cross-lingual parsing as a special case of cross-domain technically.

The importance of joint models is decreasing.
By using neural networks, global features across different tasks can be directly captured by sophisticated neural structures such as deep LSTM and self-attention,
and on the other hand, we can build one share encoder across tasks to reduce the influence of error propagation.
For parser application, which might be regarded as the reverse direction of joint models,
neural network encoders can lead to highly effective and elegant universal representations with syntactic and semantic information.
Also, all current state-of-the-art methods still require a comprehensive and fair comparison.

Finally, treebank development is the major source of the advances of syntactic and semantic parsing,
which might be the most difficult and highly valuable job.
In particular, the semantic knowledge of one sentence can have several different views.
Comprehensive annotations require extremely-high costs.
How to effectively perform treebank annotation is one task deserved investigation.

For future directions, there is still a lot of work left to be followed.
Most importantly, parsing with more complex grammars would receive increasing attention,
although this survey is no covered.
For syntactic parsing, the performances of CCG, HPSG and LFG parsing are still unsatisfactory, especially for non-English languages.
For semantic parsing, the dependency-based grammar is not enough for rich semantics,
even being relaxed with graph constraints.
Non-lexicalized nodes are necessary to express several complicated semantics.
Thus, AMR, UCCA and MRS could be promising for practical deep semantic parsing.
Based on the CFG and dependency grammars,
the cross-domain and cross-lingual settings are deserved to be concerned,
which can be further unified.
Lightly-supervised or zero-shot models might be practical solutions.
For the joint models as well as parser applications,
multi-task-learning and pretraining might become more popular architectures for adaption.

\section*{Acknowledgments}
This work is supported by National Natural Science Foundation of China (NSFC) grants 61602160 and 61672211.

\begingroup
    \setlength{\bibsep}{0pt}
    \setstretch{1}
    \bibliographystyle{unsrtnat}
    \bibliography{survey-syntax-semantic}
\endgroup
\renewcommand{\thesection}{Appendix}
\begin{appendix}

\end{appendix}

\end{multicols}

\end{CJK}
\end{document}